\documentclass[sigconf]{acmart}
\DeclareUnicodeCharacter{1EBF}{\'{e}}

\usepackage{multirow}
\usepackage{booktabs}
\usepackage{graphicx} 
\usepackage{subcaption}
\usepackage{amsfonts}       
\usepackage{nicefrac}       
\usepackage{microtype}      
\usepackage{colortbl}
\usepackage{caption}
\usepackage{multirow}
\usepackage{float}
\usepackage{pifont}
\usepackage{arydshln}
\usepackage{fontawesome}
\usepackage{svg}

\usepackage{tcolorbox}  
\usepackage{adjustbox}  
\usepackage{geometry}   
\usepackage{arydshln}
\usepackage{enumitem}

\def\mdname{UniME}
\definecolor{kcgreen}{rgb}{0.1, 0.7, 0.2}
\definecolor{kcred}{rgb}{139, 0, 0}
\definecolor{green_ours}{HTML}{D8ECD1}
\definecolor{color1}{RGB}{157,27,229}
\definecolor{color2}{RGB}{127,220,244}
\definecolor{mygreen}{rgb}{0.1, 0.7, 0.2}
\definecolor{myred}{rgb}{139, 0, 0}
\definecolor{myorange}{rgb}{0.93, 0.51, 0.18}
\definecolor{LightCyan}{rgb}{0.96,0.96,0.96}
\definecolor{lightblue}{RGB}{240,240,240}

\copyrightyear{2025}
\acmYear{2025}
\setcopyright{cc}
\setcctype{by}
\acmConference[MM '25]{Proceedings of the 33rd ACM International Conference on Multimedia}{October 27--31, 2025}{Dublin, Ireland}
\acmBooktitle{Proceedings of the 33rd ACM International Conference on Multimedia (MM '25), October 27--31, 2025, Dublin, Ireland}\acmDOI{10.1145/3746027.3754845}
\acmISBN{979-8-4007-2035-2/2025/10}
\begin{document}

\title{Breaking the Modality Barrier: Universal Embedding Learning with Multimodal LLMs}

\author{Tiancheng Gu$^\text{\ding{170}}$\textsuperscript{*}, 
Kaicheng Yang$^{\text{\ding{171}}}$\textsuperscript{*}, Ziyong Feng$^{\text{\ding{171}}}$, Xingjun Wang$^{\clubsuit}$, Yanzhao Zhang$^{\clubsuit}$, \\ Dingkun Long$^{\clubsuit}$, Yingda Chen$^{\clubsuit}$,
Weidong Cai$^{\text{\ding{170}}}$,
Jiankang Deng$^{\blacklozenge}$\textsuperscript{$\dagger$}\\
$^{\text{\ding{170}}}$The University of Sydney $^{\text{\ding{171}}}$DeepGlint $^{\clubsuit}$Tongyi Lab, Alibaba Group$^{\blacklozenge}$Imperial College London \\
\texttt\tiny{tigu8498@uni.sydney.edu.au, kaichengyang@deepglint.com}
\\
{\faGlobe\ \href{https://garygutc.github.io/UniME}{\textcolor{magenta}{\textit{Project Page}}}}  \hspace{2cm}
{\faGithub\ \href{https://github.com/deepglint/UniME}{\textcolor{magenta}{\textit{Code}}}}
}
\thanks{\textsuperscript{*} Equal Contribution \\
\textsuperscript{$\dagger$} Corresponding Author}

\renewcommand{\shortauthors}{Tiancheng Gu, Kaicheng Yang et al.}

\begin{abstract}
The Contrastive Language-Image Pre-training (CLIP) framework has become a widely used approach for multimodal representation learning, particularly in image-text retrieval and clustering. However, its efficacy is constrained by three key limitations: (1) text token truncation, (2) isolated image-text encoding, and (3) deficient compositionality due to bag-of-words behavior. While recent Multimodal Large Language Models~(MLLMs) have demonstrated significant advances in generalized vision-language understanding, their potential for learning transferable multimodal representations remains underexplored.
In this work, we present \textbf{UniME}~(\textbf{Uni}versal \textbf{M}ultimodal \textbf{E}mbedding), a novel two-stage framework that leverages MLLMs to learn discriminative representations for diverse downstream tasks.
In the first stage, we perform textual discriminative knowledge distillation from a powerful LLM-based teacher model to enhance the embedding capability of the MLLM’s language component.
In the second stage, we introduce hard negative enhanced instruction tuning to further advance discriminative representation learning. Specifically, we initially mitigate false negative contamination and then sample multiple hard negatives per instance within each batch, forcing the model to focus on challenging samples. This approach not only improves discriminative power but also enhances instruction-following ability in downstream tasks.
We conduct extensive experiments on the MMEB benchmark and multiple retrieval tasks, including short\&long caption retrieval and compositional retrieval. Results demonstrate that UniME achieves consistent performance improvement across all tasks, exhibiting superior discriminative and compositional capabilities.
\end{abstract}

\begin{teaserfigure}
    \includegraphics[width=0.98\linewidth]{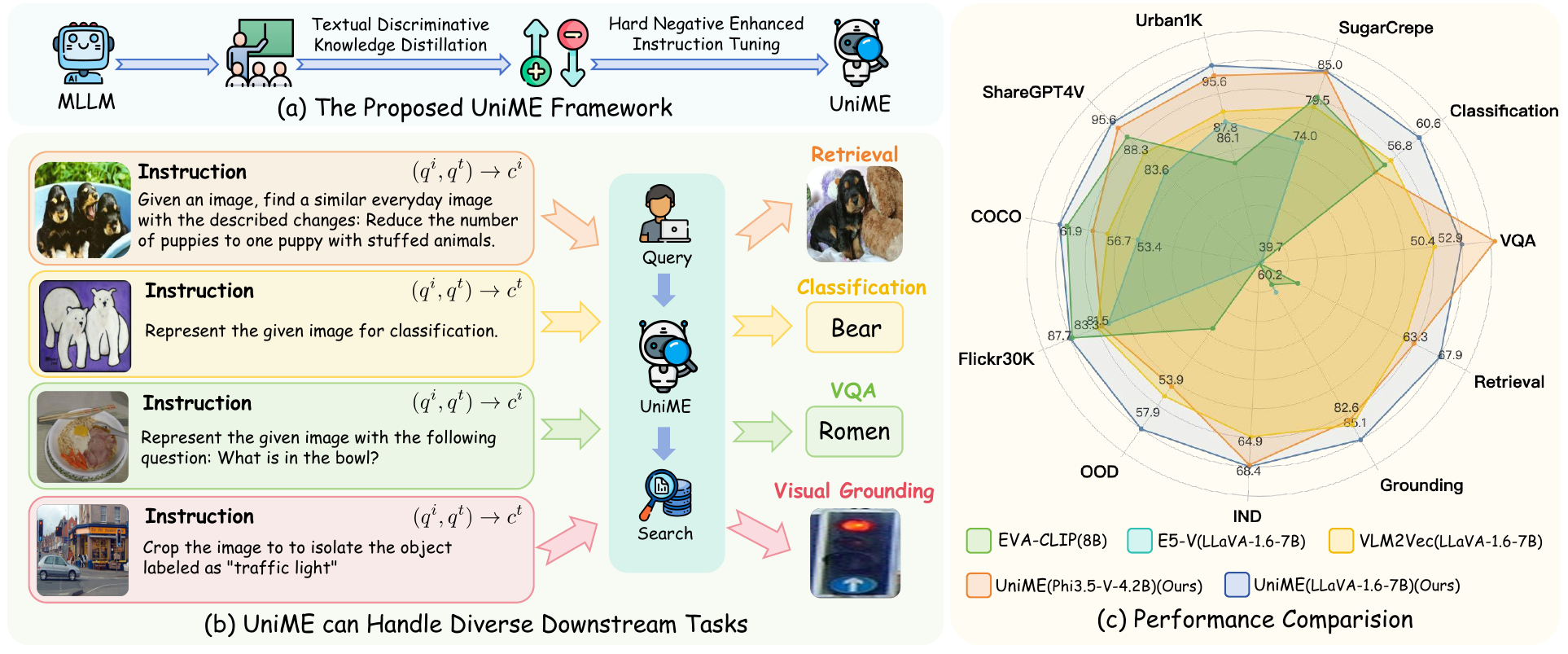}
    \vspace{-3mm}
    \caption{The UniME framework incorporates textual discriminative knowledge distillation and hard negative enhanced instruction tuning stages to learn discriminative representations for diverse downstream tasks. Our framework achieves state-of-the-art performance on both the MMEB benchmark and multiple retrieval tasks.}
    \Description[<short description>]{<long description>}
    \label{fig:introduction}
\end{teaserfigure}


\maketitle

\section{Introduction}
\label{sec:intro}
\sloppy
Modern AI applications increasingly rely on multimodal embeddings to process diverse data types, powering essential tasks like image-text retrieval~\cite{baldrati2023zero, tang2025missing}, Retrieval Augmented Generation~(RAG)~\cite{jiang2023active, cong-etal-2023-universal}, and Visual Question Answering (VQA)~\cite{garderes2020conceptbert,colpali,chun2021probabilistic}. As a seminal model, CLIP~\cite{CLIP} demonstrates notable text-image retrieval performance via cross-modal contrastive supervision using large-scale web-collected image-text pairs. However, despite its widespread use, CLIP presents notable limitations. Firstly, it restricts text token length to 77, hindering its ability to process detailed descriptions and limiting its utility in cross-modal retrieval tasks that require extensive contextual information~\cite{longclip,flame,llm2clip}. Moreover, CLIP employs a dual-encoder architecture that processes images and text separately, which compromises its effectiveness in complex tasks such as instruction-following multimodal retrieval~\cite{VLM2Vec, LamRA, wei2024uniir}. Additionally, CLIP exhibits limited advanced language understanding, struggles with compositionality, and tends to display bag-of-words behavior~\cite{negclip,tschannen2023image}.

The success of Large Language Models (LLMs)~\cite{llama,llama2,llama3,qwen,qwen2.5} has motivated researchers to adapt LLMs to understand multimodal inputs. Multimodal Large Language Models~(MLLMs) as a key component in the construction of general-purpose AI assistants have demonstrated remarkable progress~\cite{llava,llava1.5}. For example, Qwen2-VL~\cite{qwen2vl} innovates beyond fixed-resolution visual processing, achieving robust performance across diverse image resolutions and aspect ratios. Similarly, LLaVA-OneVision~\cite{llava-ov} introduces a unified modeling approach that enables effective task transfer across scenarios while maintaining architectural simplicity. While these MLLMs show impressive vision-language reasoning capabilities, these MLLMs are inherently constrained by their autoregressive next-token prediction objective, which limits their effectiveness in learning multimodal representations compared to contrastive methods such as CLIP~\cite{E5V,VLM2Vec}.

Recent advances in LLM-based models have demonstrated substantial progress on the MTEB benchmark~\cite{mteb}. Inspired by these developments~\cite{nvembed,llm2vec}, researchers are now actively investigating MLLMs for unified multimodal representation learning. E5-V~\cite{E5V} proposes an unimodal contrastive learning approach that trains the language component of MLLM on sentence pairs to bridge cross-modal representation disparities. However, this method encounters two primary constraints: (1) constraints arising from the limited scale and diversity of training data~\cite{discriminative}; (2) inherent challenges caused by the MLLM's causal attention mechanism, which fundamentally restricts its ability to learn complex contextual representations~\cite{ullme,croc,show-o}. These factors collectively constrain the model's full embedding potential. VLM2Vec~\cite{VLM2Vec} introduces the Massive Multimodal Embedding Benchmark (MMEB), comprising 36 datasets across 4 meta-tasks, and develops a contrastive framework that converts state-of-the-art vision-language models into embedding models through MMEB training. Nevertheless, the existence of false negative samples in the batch significantly complicates the discrimination of hard negative pairs when using the standard InfoNCE loss.

To overcome these challenges, we present \textbf{UniME} (\textbf{Uni}versal \textbf{M}ultimodal \textbf{E}mbedding), a novel two-stage framework that empowers multimodal large language models~(as shown in Figure~\ref{fig:introduction}) to learn universal representations for diverse downstream vision-language tasks.
In the first textual discriminative knowledge distillation stage, we leverage a powerful LLM-based teacher model to enhance the embedding capabilities of MLLM's language component.
In the second stage of hard negative enhanced instruction tuning, we first eliminate false negative contamination, then implement a hard negative sampling strategy that selects multiple challenging negatives per instance within each batch. This approach forces the model to focus on challenging negative samples, thereby learning more discriminative multimodal representations while also improving instruction-following ability in downstream tasks. We evaluate our approach comprehensively on the MMEB benchmark and multiple retrieval tasks, including both short\&long caption retrieval and compositional retrieval. Experimental results demonstrate that UniME achieves significant performance improvement across all tasks, exhibiting both robust discriminative power and superior compositional understanding. The main contributions of this paper are summarized as follows:
\begin{itemize}[leftmargin=*]
    \item We present \textbf{UniME} (\textbf{Uni}versal \textbf{M}ultimodal \textbf{E}mbedding), \textbf{a novel two-stage framework} that empowers Multimodal Large Language Models (MLLMs) to learn universal representations for diverse downstream tasks.
    \item We propose \textbf{textual discriminative knowledge distillation}, leveraging a powerful LLM-based teacher model to enhance the embedding capability of the MLLM’s language component.
    \item We introduce \textbf{hard negative enhanced instruction tuning} to further advance discriminative representation learning through false negative filtering and hard negative sampling.
    \item We conduct \textbf{extensive experiments} on the MMEB benchmark and multiple retrieval tasks, including both short\&long caption retrieval and compositional retrieval. Results show that UniME demonstrates robust discriminative and compositional capabilities, achieving notable performance improvements across all evaluated tasks.
 \end{itemize}
\section{Related work}
\label{sec:Related work}

\begin{figure*}[t!]
    \centering
    \includegraphics[width=\linewidth]{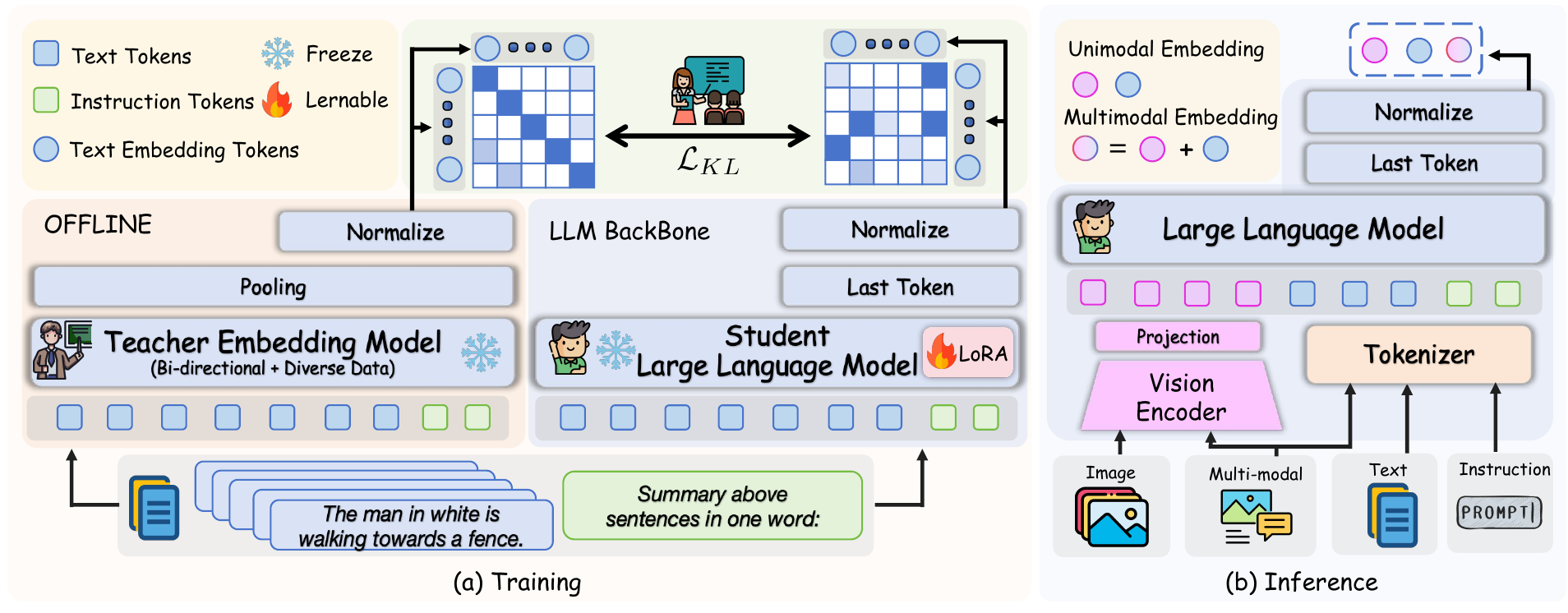}
    \vspace{-7mm}
    \caption{The framework of the Textual Discriminative Knowledge Distillation stage. We leverage the state-of-the-art LLM-based embedding model to enhance the discriminative capabilities of the MLLM's language component.}
    \Description[<short description>]{<long description>}
    \vspace{-3mm}
    \label{fig:method_distillation}
\end{figure*}

\subsection{Multimodal Large Language Models}
Multimodal Large Language Models (MLLMs) extend LLMs to process and integrate cross-modal information~\cite{bin2024gallerygpt, wang2024large, wang2024large, jiang2024hal}. A seminal development in this field is LLaVA~\cite{llava}, which encodes images into token representations using vision encoders like CLIP~\cite{CLIP, EVA_CLIP, yang2023alip, gu2024rwkv, yang2025clip, hu2025decoupled} and projects them into the LLM's token space. Following this breakthrough, numerous MLLM variants~\cite{kosmos,vila,llavagrounding,minigpt4} have demonstrated remarkable performance in multimodal comprehension and reasoning tasks. CogVLM~\cite{wang2023cogvlm} incorporates a trainable visual expert module within the attention and feed-forward network layers of the language model, yielding substantial performance enhancements across 17 canonical cross-modal benchmarks. LLaVA-1.6~\cite{llavanext} introduces the ``AnyRes'' technique to process variable high-resolution images, significantly improving fine-grained visual understanding. Phi3.5-Vision~\cite{phi3}, a 4.2-billion parameter model evolved from phi-3.5-mini, exhibits strong reasoning capabilities for both single and multi-image text prompts. While these advancements, the autoregressive next-token prediction objective of MLLM inherently constrains its capacity to learn efficient multimodal representations.

\subsection{LLMs for Representation Learning}
As large language models increasingly exhibit remarkable proficiency in natural language processing, recent research has pivoted towards harnessing decoder-only architectures for effective representation learning~\cite{ma2024fine,nvembed,llm2vec, shin2025generative, didolkar2025ctrl}. Previous work has adapted the prompt-based representation method for autoregressive models, enabling Large Language Models (LLMs) to perform in-context learning and scale to various model sizes. LLM2Vec~\cite{llm2vec} transforming pre-trained decoder-only LLMs into versatile text encoders by incorporating three principal advancements: bidirectional attention mechanisms, masked next-token prediction, and unsupervised contrastive alignment. Concurrently, NV-Embed~\cite{nvembed} introduces a latent attention layer and eliminates the causal attention mask during contrastive training, substantially enhancing the efficiency of embeddings generated from decoder-only LLMs. While these approaches show promising embedding performance, their exclusive focus on text-only inputs fails to meet the growing demands of multimodal applications.
 
\subsection{Multimodal Representation Learning}
CLIP~\cite{CLIP} demonstrates superior image-text retrieval capabilities through large-scale cross-modal contrastive learning, but it suffers from three inherent constraints: (1) 77-token text truncation limits fine-grained semantic alignment~\cite{longclip,flame,llm2clip}; (2) Disjoint dual-encoder architecture impedes cross-modal fusion for instruction-sensitive tasks~\cite{VLM2Vec, LamRA, wei2024uniir}; (3) Primitive language modeling induces bag-of-words representations~\cite{negclip,tschannen2023image}. Recent research addresses these limitations through two complementary approaches.
Firstly, MagicLens~\cite{magiclens} employs lightweight dual encoders to enable relation-aware image retrieval guided by textual instructions. Secondly, MLLM-based approaches emerge for robust representation learning. E5-V~\cite{E5V} proposes a single-modality training approach that significantly outperforms traditional multimodal training on image-text pairs while reducing training costs. VLM2Vec~\cite{VLM2Vec} presents a contrastive training framework that can handle any combination of images and text, as well as high-resolution images and long text inputs. Despite these improvements, current methods still face challenges in effectively discriminating hard negative samples during retrieval.

\section{Method}
\label{sec:method}

\begin{figure*}[t!]
    \centering
    \includegraphics[width=\linewidth]{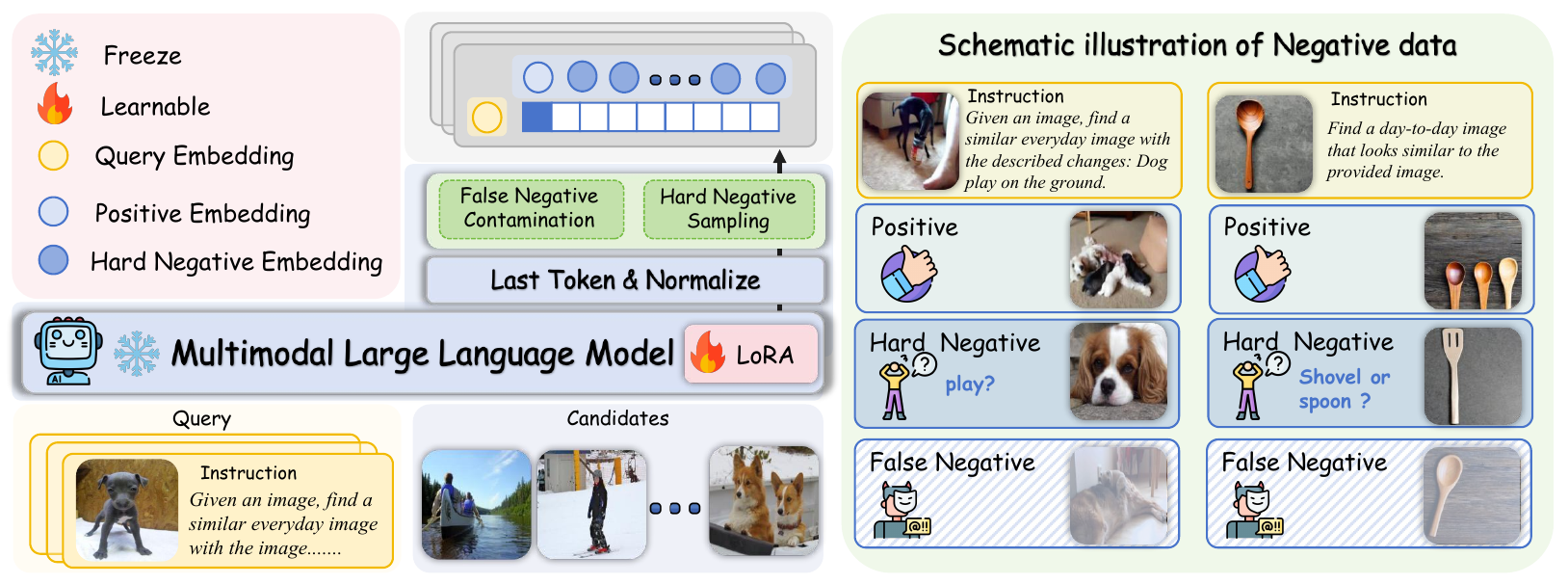}
    \vspace{-7mm}
    \caption{The framework of the Hard Negative Enhanced Instruction Tuning stage. We further improve the discriminative capabilities of the MLLM through false negative filtering and hard negative sampling.}
    \Description[<short description>]{<long description>}
    \vspace{-3mm}
    \label{fig:method_stage2_alignment}
\end{figure*}

This section first establishes the preliminary definitions, including task formulation and feature extraction (Section~\ref{subsec:problem formulation}). Then we start to introduce our proposed novel two-stage framework \mdname. We elaborate on the first textual discriminative knowledge distillation stage in Section~\ref{subsec:stage1}. After that, we present the second hard negative enhanced instruction tuning stage in Section~\ref{subsec:stage2}. Finally, we introduce the training recipe in Section~\ref{sub:training_recipe}.

\subsection{Preliminary}
\label{subsec:problem formulation}

\subsubsection{Task Definition.} 
\sloppy
To address the limitations of dual-tower encoder structures in acquiring unified multimodal representations~\cite{CLIP}, we employ Multimodal Large Language Models~(MLLMs) with robust multimodal understanding to learn universal multimodal embeddings. Specifically, we feed both the query and candidate data into the MLLM using customized prompts to extract their respective embeddings. We then calculate the similarity~($\Theta$) between the query~($\boldsymbol{q}$) and candidates~($\boldsymbol{c}$), followed by ranking and selecting the most relevant pairs~($\boldsymbol{P}$). This procedure is formalized as follows:
\begin{equation}
\begin{aligned}
    & \boldsymbol{P} = \text{Rank}(\Theta(\phi(\boldsymbol{q}), \phi(\boldsymbol{c}))),
\end{aligned}
\end{equation}
where $\phi$ denotes the MLLM employed to extract respective embeddings. Notably, both the query and candidate may be either unimodal (text or image only) or multimodal (interleaved image-text).

\subsubsection{Feature Extraction by MLLM.}
\sloppy
Unlike the dual-tower structure of CLIP, MLLM incorporates three essential components: a vision tower, a projection layer, and an LLM backbone. This unified structure supports flexible processing of both unimodal (image or text) and multimodal (interleaved image-text) inputs, enabling diverse task execution within a single framework. In this work, we present a novel two-stage framework that enables MLLMs to learn universal multimodal embedding for diverse downstream tasks.
In the first textual discrimination knowledge distillation stage, we follow the previous work~\cite{E5V} and employ the prompt: \textit{``<Text> Summary above sentences in one word: $\backslash$n''} to guide the LLM to compress textual information into a single embedding at the last token. In the second hard negative enhanced instruction tuning stage, we utilize task-specific prompts from VLM2Vec~\cite{VLM2Vec} to adapt \mdname\ fit diverse downstream tasks such as: \textit{"<Image> Represent the given image with the following question: <Text>"} for Visual Question Answering~(VQA) tasks and \textit{"<Image> Find a caption for the news in the given photo."} for retrieval tasks.

\subsection{Textual Discriminative Knowledge Distillation}
\label{subsec:stage1}
\sloppy
\subsubsection{Training.} 
Inspired by previous research~\cite{E5V}, we enhance the LLM backbone of the MLLM to improve its overall embedding capabilities. The autoregressive decoder architecture of LLMs, constrained by a causal masking mechanism, inherently restricts their discriminative capacity and poses challenges in effectively distinguishing between diverse items. To address this limitation, we introduce discriminative textual knowledge distillation~(as shown in Figure~\ref{fig:method_distillation}), transferring knowledge from the state-of-the-art LLM-based embedding model NV-Embed V2~\cite{nvembed}, which employs multiple diverse datasets and removes the causal attention mask during contrastive training. Specifically, we first decouple the LLM component from the MLLM architecture and process text-only inputs using the embedding prompt: \textit{``<Text> Summary above sentences in one word: $\backslash$n''}. Then we obtain the normalized student text embeddings $e_s \in \mathbb{R}^{n\times d}$ and teacher text embeddings~(which is extracted offline) $e_t \in \mathbb{R}^{n\times d}$ from the hidden state of the final token, where $n$ is the batch size, $d$ is the dimension of the embeddings. Subsequently, we implement discriminative distribution alignment by minimizing the Kullback-Leibler (KL) divergence~\cite{KL_diver} between the embeddings from the teacher model and student model:
\begin{equation}
    \mathcal{L}_{KL} = \sum_{i=1}^{n}\mathrm{KL}
    \left( 
        \frac{\exp\left(e_{s_i}^\top e_{s_i} / \tau\right)}
             {\sum\limits_{j=1}^n \exp\left(e_{s_j}^\top e_{s_i} / \tau\right)} 
        \,\Bigg\Vert\, 
        \frac{\exp\left(e_{t_i}^\top e_{t_i} / \tau\right)}
             {\sum\limits_{j=1}^n \exp\left(e_{t_j}^\top e_{t_i} / \tau\right)} 
    \right),
\end{equation}
where $\tau$ is the temperature hyper-parameter used to soften distribution representation. By distilling the relationships between different samples within a batch, our method demonstrates enhanced efficiency compared to direct contrastive learning under identical data and training conditions and achieves significant performance improvements in downstream tasks.

\subsubsection{Inference.} During the training phase, our approach exclusively utilizes text-only inputs and optimizes solely the language model component within the multimodal language model architecture, while maintaining other parameters frozen. For inference, we restore the original vision encoder and projection layer to enable multimodal processing. For unimodal inputs (text or image), we use modality-specific standardized prompts. For interleaved image-text inputs, we process each modality independently with its corresponding prompt and aggregate the embeddings through element-wise summation to produce the final multimodal representation.

\begin{table*}[t]
\centering
\caption{Results on the MMEB benchmark. IND represents the in-distribution dataset, and OOD represents the out-of-distribution dataset. The reported scores are the average Precision@1 over the corresponding datasets. The best results are marked in bold. $^\dagger$: UniME with textual discrimination distillation only. $^\ddagger$: UniME with both textual discrimination distillation and hard negative enhanced instruction tuning.}
\vspace{-2mm}
\setlength\tabcolsep{7pt}
\renewcommand\arraystretch{1.2}
\small
\fontsize{8pt}{8pt}\selectfont
\resizebox{\textwidth}{!}{
\begin{tabular}{lcccccccc}
\toprule
\multicolumn{1}{l}{\multirow{2}{*}{\textbf{Models}}} & \multicolumn{1}{c}{\multirow{2}{*}{\textbf{\#Parameters}}} & \multicolumn{4}{c}{\textbf{Per Meta-Task Score}} & \multicolumn{3}{c}{\textbf{Average Score}}    \\ \cmidrule(lr){3-6} \cmidrule(lr){7-9} 
\multicolumn{1}{c}{} & \multicolumn{1}{c}{} &  Classification & VQA & Retrieval & Grounding & IND & OOD & Overall \\ \midrule
\# of Datasets $\rightarrow$  & & 10       & 10            & 12            & 4             & 20            & 16            & 36            \\ \midrule
\multicolumn{9}{c}{\textcolor{black}{\textbf{\textit{Zero-shot on MMEB}}}}  \\ \midrule

CLIP(ViT-L)~\cite{VLM2Vec}   & 0.4B & 42.8 & 9.1 & 53.0 & 51.8 & 37.1 & 38.7 & 39.2 \\
OpenCLIP(ViT-L)~\cite{CLIP}   &  0.4B & 41.5 & 6.9 & 44.6 & 53.5 & 32.8 & 36.0 & 36.6 \\
Magiclens(ViT-L)~\cite{magiclens} & 0.4B & 38.8 & 8.3 & 35.4 & 26.0 & 31.0 & 23.7 & 27.1 \\
SigLIP(So/14)~\cite{siglip} & 0.9B & 40.3 & 8.4 & 31.6 & 59.5 & 32.3 & 38.0 & 35.0 \\
BLIP2(ViT-L)~\cite{blip2} & 1.2B & 27.0 & 4.2 & 33.9 & 47.0 & 25.3 & 25.1 & 28.0 \\
CLIP(ViT-BigG/14)~\cite{CLIP_bigG14} & 2.5B & 52.3 & 14.0 & 50.5 & 60.3 & 38.9 & 45.8 & 44.3 \\
EVA-CLIP~\cite{EVA_CLIP_18B} & 8B &  56.0 & 10.4 & 49.2 & 58.9 & 38.1 & 45.6 & 43.7   \\
\hdashline
E5-V(Phi3.5-V)~\cite{E5V} & 4.2B & 39.1 & 9.6 & 38.0 & 57.6 & 33.1 & 31.9 & 36.1 \\
E5-V(LLaVA-1.6)~\cite{E5V} & 7B & 39.7 & 10.8 & 39.4 & 60.2 & 34.2 & 33.4 & 37.5 \\
\rowcolor[HTML]{EDEDED}
\rowcolor[HTML]{EDEDED}
\mdname$^\dagger$(Phi3.5-V) & 4.2B & 42.5(\textcolor{kcgreen}{+3.4}) & 18.3(\textcolor{kcgreen}{+8.7}) & 40.5(\textcolor{kcgreen}{+2.5}) & 59.9(\textcolor{kcgreen}{+2.3}) & 36.0(\textcolor{kcgreen}{+2.9}) & 38.3(\textcolor{kcgreen}{+6.4}) & 40.3(\textcolor{kcgreen}{+4.2}) \\
\rowcolor[HTML]{EDEDED}
\mdname$^\dagger$(LLaVA-1.6) & 7B & 43.0(\textcolor{kcgreen}{+3.3}) & 17.7(\textcolor{kcgreen}{+6.9}) & 42.5(\textcolor{kcgreen}{+3.1}) & 63.2(\textcolor{kcgreen}{+3.0}) & 37.6(\textcolor{kcgreen}{+3.4}) & 38.6(\textcolor{kcgreen}{+5.2}) & 41.6(\textcolor{kcgreen}{+4.1}) \\ 
\midrule

\multicolumn{9}{c}{\textcolor{black}{\textbf{\textit{Fine-tuning on MMEB}}}}  \\ 
\midrule

CLIP(ViT-L)~\cite{VLM2Vec}  & 0.4B & 55.2 & 19.7 & 53.2 & 62.2 & 47.6 & 42.8 & 47.6 \\
\hdashline
VLM2Vec(Phi3.5-V)~\cite{VLM2Vec} & 4.2B & 54.8 & 54.9 & 62.3 & 79.5 & 66.5 & 52.0 & 62.9 \\
VLM2Vec(LLaVA-1.6)~\cite{VLM2Vec} & 7B & 56.8 & 50.4 & 63.3 & 82.6 & 64.9 & 53.9 & 63.3  \\
\rowcolor[HTML]{EDEDED}
\rowcolor[HTML]{EDEDED}
\mdname$^\ddagger$(Phi3.5-V) & 4.2B & 54.8(\textcolor{kcgreen}{+0.0}) & \textbf{55.9}(\textcolor{kcgreen}{+1.0}) & 64.5(\textcolor{kcgreen}{+2.2}) & 81.8(\textcolor{kcgreen}{+2.3}) & 68.2(\textcolor{kcgreen}{+1.7}) & 52.7(\textcolor{kcgreen}{+0.7}) & 64.2(\textcolor{kcgreen}{+1.3}) \\
\rowcolor[HTML]{EDEDED}
\mdname$^\ddagger$(LLaVA-1.6) & 7B & 60.6(\textcolor{kcgreen}{+3.8}) & 52.9(\textcolor{kcgreen}{+2.5}) & 67.9(\textcolor{kcgreen}{+4.6}) & 85.1(\textcolor{kcgreen}{+2.5}) & 68.4(\textcolor{kcgreen}{+3.5}) & 57.9(\textcolor{kcgreen}{+4.0}) & 66.6(\textcolor{kcgreen}{+3.3}) \\
\rowcolor[HTML]{EDEDED}
\mdname$^\ddagger$(LLaVA-OneVision) & 7B & \textbf{66.8} & \textbf{66.6} & \textbf{70.6} & \textbf{90.9} & \textbf{74.6} & \textbf{65.8} & \textbf{70.7} \\

\bottomrule
\end{tabular}}
\vspace{-3mm}
\label{tab:main_results}
\end{table*}

\subsection{Hard Negative Enhanced Instruction Tuning}
\label{subsec:stage2}
After textual discriminative knowledge distillation, \mdname\ develops preliminary discriminative capabilities but exhibits limited visual sensitivity. This insensitivity results in deviations in image-text alignment and limits the discriminative performance, despite the MLLM's extensive pretraining on large-scale datasets. Moreover, the generic instruction prompts used in the first stage hinder \mdname's effectiveness in complex retrieval tasks. To address these limitations, we introduce an additional hard negative enhanced instruction tuning stage~(as shown in Figure~\ref{fig:method_stage2_alignment}), which aims to: (1) further enhance discriminative capabilities, (2) improve cross-modal alignment, and (3) strengthen instruction-following ability for downstream tasks.

\subsubsection{False Negative Contamination.} 
The presence of false negatives in training batches hinders effective hard negative differentiation under the standard InfoNCE loss. As evidenced in Table~\ref{tab:alpha}, false negatives frequently appear as candidate samples. To mitigate this, we introduce a filtering mechanism based on the similarity threshold between the query and positive samples, defined as: $\alpha = cos(e_{q}, e_{c}^{+}) + \beta$, where $\beta$ is a hyper-parameter used to control the threshold margin. During training, we exclude all negative samples whose similarity to the query exceeds $\alpha$, effectively eliminating false negatives while preserving challenging hard negatives.

\subsubsection{Hard Negative Sampling}
Hard negative samples, distinct in label from the positive sample but closely embedded, are poised to offer the greatest utility through the provision of substantial gradient information in the context of contrastive learning. In contrast, easy negatives yield negligible gradients and contribute minimally to the learning process. Drawing inspiration from prior research \cite{karpukhin2020dense, robinsoncontrastive, kalantidis2020hard}, we propose a hard negative sampling strategy to optimize both training efficiency and discriminative performance. The textual knowledge distillation stage equips \mdname\ with preliminary discriminative ability to autonomously identify hard negatives for each query. Building on this capability, we sample $k$ corresponding hard negative  $e^{-}_{c}$ within each training batch as follows:
\begin{equation}
\begin{aligned}
    e^{-}_{c} = \text{Rank}_{k}(cos(e_{q}, e_{c})), \text{where}\ e_{c} \notin \left \{ e^{+}_{c}, e^{*}_{c}\right\},
\end{aligned}
\end{equation}
where $e^{*}_{c}$ and $e^{+}_{c}$ denote filtered false negative candidates and positive candidates respectively, $e_{q}$ is the query embedding, and $e_{c}$ represents all candidate embeddings. The function $cos(\cdot)$ computes pairwise similarity scores, and $\text{Rank}_{k}$ selects the top-$k$ highest-scoring candidates as hard negatives. To maintain batch consistency when fewer than $k$ hard negatives are obtained~(which occurs rarely), we duplicate existing hard negatives to preserve the fixed size $k$.

\subsubsection{Training Objective.}
After obtaining the embedding of the queries~($e_{q}$), positive candidates~($e_c^+$) and hard negative candidates~($e_c^-$), we utilize the Noise Contrastive Estimation~(InfoNCE) loss~\cite{oord2018representation} over the in-batch sampled hard negatives as follows:
\begin{equation}
\begin{aligned}
    &\mathcal{L} = -log \frac{exp(cos(e_{q}, e_{c}^{+})/\tau)}{exp(cos(e_{q}, e_{c}^{+})/\tau) + \sum_{i=1}^{k}exp(cos(e_{q}, e_{c_{i}^{-}})/\tau)}, \\
\end{aligned}
\end{equation}
where $k$ denotes the set of all hard negatives, and $\tau$ is a temperature hyper-parameter.

\begin{table*}[!ht]
  \small
    \centering
      \caption{Results of zero-shot text-image retrieval on short caption datasets (Flickr30K and MS-COCO), long caption datasets (ShareGPT4V and Urban1K) and compositional benchmark (SugarCrepe). The reported scores are the average Recall@1 over the corresponding datasets. The best results are marked in bold. $^\dagger$: UniME with textual discrimination distillation only. $^\ddagger$: UniME with both textual discrimination distillation and hard negative enhanced instruction tuning.}
      \vspace{-2mm}
      \label{tab:zero-shot-retrieval}
     \resizebox{\textwidth}{!}{
     \begin{tabular}{lcccccccccccc}
        \toprule
         \multicolumn{1}{c}{\multirow{4}{*}{\textbf{Models}}} & \multicolumn{1}{c}{\multirow{4}{*}{\textbf{\#Parameters}}} & \multicolumn{4}{c}{\textbf{Short Caption Retrieval}} & \multicolumn{4}{c}{\textbf{Long Caption Retrieval}} & \multicolumn{3}{c}{\textbf{Compositional Retrieval}} \\
        \cmidrule(lr){3-6} \cmidrule(lr){7-10} \cmidrule(lr){11-13}
         && \multicolumn{2}{c}{Flickr30K} & \multicolumn{2}{c}{COCO} & \multicolumn{2}{c}{ShareGPT4V} & \multicolumn{2}{c}{Urban1K} & \multicolumn{3}{c}{SugarCrepe} \\
        \cmidrule(lr){3-4} \cmidrule(lr){5-6} \cmidrule(lr){7-8} \cmidrule(lr){9-10} \cmidrule(lr){11-13}
        && $q^t\rightarrow c^i$ & $q^i\rightarrow c^t$ & $q^t\rightarrow c^i$ & $q^i\rightarrow c^t$ & $q^t\rightarrow c^i$ & $q^i\rightarrow c^t$ & $q^t\rightarrow c^i$ & $q^i\rightarrow c^t$  & Replace & Swap & Add \\
        \midrule
        OpenCLIP(ViT-L)~\cite{CLIP}   & 0.4B & 67.3 & 87.2 & 37.0  & 58.1 & 81.8 & 84.0 & 47.0 & 47.0 & 79.5 & 62.7 & 74.9  \\ 
        CLIP(ViT-BigG/14)~\cite{CLIP_bigG14}  & 2.5B & 79.5 & 92.9 & 51.3  & 67.3 & 90.1 & 93.6 & 77.8 & 80.7 & 86.5 & 68.9 & 88.4 \\
        EVA-CLIP~\cite{EVA_CLIP_18B}  & 8B & 80.3 & \textbf{94.5} &  52.0 & 70.1 & 93.1 & 91.2 & 80.4 & 77.8 & 85.9 & 70.3 & 86.7  \\
        \hdashline
        E5-V(Phi3.5-V)~\cite{E5V} & 4.2B & 72.2 & 79.6 & 44.7  & 53.4 & 86.0 & 88.5 & 83.8 & 83.6 & 88.2 & 66.6 & 75.3  \\
        E5-V(LLaVA-1.6)~\cite{E5V} & 7B & 77.3 & 85.7 & 49.1  & 57.6 & 85.1 & 82.1 & 88.9 & 83.2 & 86.3 & 68.7 & 66.9  \\
        \rowcolor[HTML]{EDEDED}
        \mdname$^\dagger$(Phi3.5-V) & 4.2B & 72.0(\textcolor{kcred}{-0.2}) & 80.6(\textcolor{kcgreen}{+1.0}) & 44.9(\textcolor{kcgreen}{+0.2}) & 57.2(\textcolor{kcgreen}{+0.8}) & 86.8(\textcolor{kcgreen}{+3.8}) & 92.3(\textcolor{kcgreen}{+1.3}) & 85.1(\textcolor{kcgreen}{+2.3}) & 86.9(\textcolor{kcgreen}{+3.3}) & \textbf{90.2}(\textcolor{kcgreen}{+2.0}) & 67.6(\textcolor{kcgreen}{+1.0}) & 91.2(\textcolor{kcgreen}{+15.9}) \\
        \rowcolor[HTML]{EDEDED}
        \mdname$^\dagger$(LLaVA-1.6) & 7B & 77.2(\textcolor{kcred}{-0.1}) & 84.6(\textcolor{kcred}{-1.1}) & 51.0(\textcolor{kcgreen}{+1.9}) & 56.4(\textcolor{kcred}{-1.2}) & 89.8(\textcolor{kcgreen}{+4.7}) & 86.9(\textcolor{kcgreen}{+4.8}) & 91.3(\textcolor{kcgreen}{+2.4}) & 82.4(\textcolor{kcred}{-0.8}) & 89.5(\textcolor{kcgreen}{+3.2}) & 64.8(\textcolor{kcred}{-3.9}) & 94.2(\textcolor{kcgreen}{+27.3}) \\ 
        \hdashline
        VLM2Vec(Phi3.5-V)~\cite{VLM2Vec} & 4.2B & 68.7 & 83.0 & 43.7  & 59.8 & 90.1 & 92.0 & 87.9 & 86.8 & 86.2 & 66.7 & 84.2  \\
        VLM2Vec(LLaVA-1.6)~\cite{VLM2Vec} & 7B & 76.0 & 90.6 & 46.8  & 66.6 & 85.8 & 90.7 & 84.7 & 90.8 & 85.8 & 66.3 & 86.5 \\
        \rowcolor[HTML]{EDEDED}
        \mdname$^\ddagger$ (Phi3.5-V) & 4.2B & 77.0(\textcolor{kcgreen}{+11.3}) & 88.2(\textcolor{kcgreen}{+5.2}) & 49.8(\textcolor{kcgreen}{+6.1}) & 66.8(\textcolor{kcgreen}{+7.0}) & 92.1(\textcolor{kcgreen}{+2.0}) & 96.4(\textcolor{kcgreen}{+4.4}) & 92.7(\textcolor{kcgreen}{+4.8}) & 95.1(\textcolor{kcgreen}{+8.3}) & 90.1(\textcolor{kcgreen}{+3.9}) & 70.9(\textcolor{kcgreen}{+4.2}) & 93.3(\textcolor{kcgreen}{+9.1}) \\
        \rowcolor[HTML]{EDEDED}
        \mdname$^\ddagger$ (LLaVA-1.6) & 7B & 81.9(\textcolor{kcgreen}{+5.9}) & 93.4(\textcolor{kcgreen}{+2.8}) & 53.7(\textcolor{kcgreen}{+6.1})  & 70.1(\textcolor{kcgreen}{+3.5}) & 93.9(\textcolor{kcgreen}{+8.1}) & \textbf{97.2}(\textcolor{kcgreen}{+6.5}) & \textbf{95.2}(\textcolor{kcgreen}{+10.5}) & \textbf{95.9}(\textcolor{kcgreen}{+5.1}) & 89.0(\textcolor{kcgreen}{+3.2}) & \textbf{71.5}(\textcolor{kcgreen}{+5.2}) & \textbf{94.4}(\textcolor{kcgreen}{+7.9})  \\
        \rowcolor[HTML]{EDEDED}
        \mdname$^\ddagger$ (LLaVA-OV) & 7B & \textbf{83.3} & 94.4 & \textbf{54.8} & \textbf{74.0} & \textbf{93.9} & 89.3 & 94.3 & 95.5 & 80.5 & 65.5 & 82.2 \\
        \bottomrule
    \end{tabular}
    }
    \vspace{-3mm}
\end{table*}

\subsection{Training Recipe}
\label{sub:training_recipe}
\sloppy
\noindent{\bf Stage1: Textual Discriminative Knowledge
Distillation.} We employ QLoRA (Quantized Low-Rank Adaptation)~\cite{QLora} for parameter-efficient fine-tuning of the large language model component. This stage exclusively utilizes text-only inputs and introduces minimal trainable parameters, typically less than 5\% of the total. The complete training procedures for Phi3.5-V and LLaVA-1.6 require approximately 1 hour and 2 hours, respectively.

\noindent{\bf Stage2: Hard Negative Enhanced Instruction Tuning.} To overcome GPU memory limitations in large-batch MLLM training, we employ a dual strategy: (1) Following VLM2Vec~\cite{VLM2Vec}, we implement GradCache~\cite{GradeCache}, a gradient caching technique that decouples backpropagation between contrastive loss computation and encoder updates; (2) we employ QLoRA~\cite{QLora} for parameter-efficient fine-tuning of all parameters within the MLLM. This combined approach facilitates effective training while ensuring manageable memory consumption.
\section{Experiments} 
\label{sec:exp}

\subsection{Implementation}
We evaluate our proposed method through extensive experiments on three different multimodal large language models: Phi3.5-Vision~\cite{phi3}, and LLaVA-1.6~\cite{llavanext}. Our implementation leverages PyTorch with DeepSpeed~\cite{deepspeed} ZeRO stage-2 optimization to enhance training efficiency and facilitate community adoption. The training of UniME is conducted on 8$\times$NVIDIA A100 (80GB) GPUs to accommodate the substantial computational demands.
In the textual discriminative knowledge distillation stage, we implement gradient accumulation with a batch size of 768, a learning rate of 5e-4, and a LoRA rank of 32. We employ NV-Embed V2~\cite{nvembed} (one of the state-of-the-art embedding model on the MTEB benchmark~\cite{mteb}) as our teacher model. The training completes within two epochs.
In the hard negative enhanced instruction turning stage, we use the low-resolution (336×336) image inputs. We increase the accumulated batch size to 1024, reduce the learning rate to 1e-4 for Phi3.5-V and 2e-5 for LLaVA-1.6, and decrease the LoRA rank to 16. We sample $k=8$ hard negatives within a batch with a similarity threshold of ($\beta$=0.1). Each model undergoes 1,000 training steps during this stage.

\subsection{Datasets and Metrics}
\subsubsection{Training.} 
Following E5-V~\cite{E5V}, we utilize the Natural Language Inference (NLI) dataset~\cite{simcse} which contains around 273k sentence pairs for the first textual discriminative knowledge distillation stage. For the hard negative enhanced instruction tuning stage, similar to the VLM2Vec~\cite{VLM2Vec}, we employ 20 in-distribution datasets from the MMEB benchmark, which cover four core multimodal tasks: classification, visual question answering, multimodal retrieval, and visual grounding. This comprehensive training corpus, incorporating both unimodal and multimodal input data, totals 662k carefully curated training pairs, ensuring robust model adaptation across diverse multimodal tasks.

\subsubsection{Evaluation.}
In this study, we evaluate \mdname\ across both in-distribution (20 test sets) and out-of-distribution (16 test sets) benchmarks from MMEB~\cite{VLM2Vec} to assess its multimodal embedding capabilities across diverse retrieval tasks. Following standard evaluation protocols~\cite{LamRA, VLM2Vec}, we report Precision, which quantifies the proportion of correct matches among the top-ranked candidates for each dataset.
To further examine the unimodal embedding performance of \mdname, we conduct experiments on multiple cross-modal retrieval tasks, including short-caption image-text retrieval (Flickr30K~\cite{flickr30k} and COCO2014~\cite{MSCOCO2014}), long-caption image-text retrieval (ShareGPT4V~\cite{sharegpt4v} and Urban1K~\cite{longclip}), and compositional retrieval (SugarCrepe~\cite{sugarcrepe}). Consistent with the MMEB benchmark, Precision serves as the primary evaluation metric across all datasets. 

\subsection{Main Results.}
\subsubsection{Multi-Modal Retrieval}
\begin{figure}[t!]
    \centering
    \includegraphics[width=\linewidth]{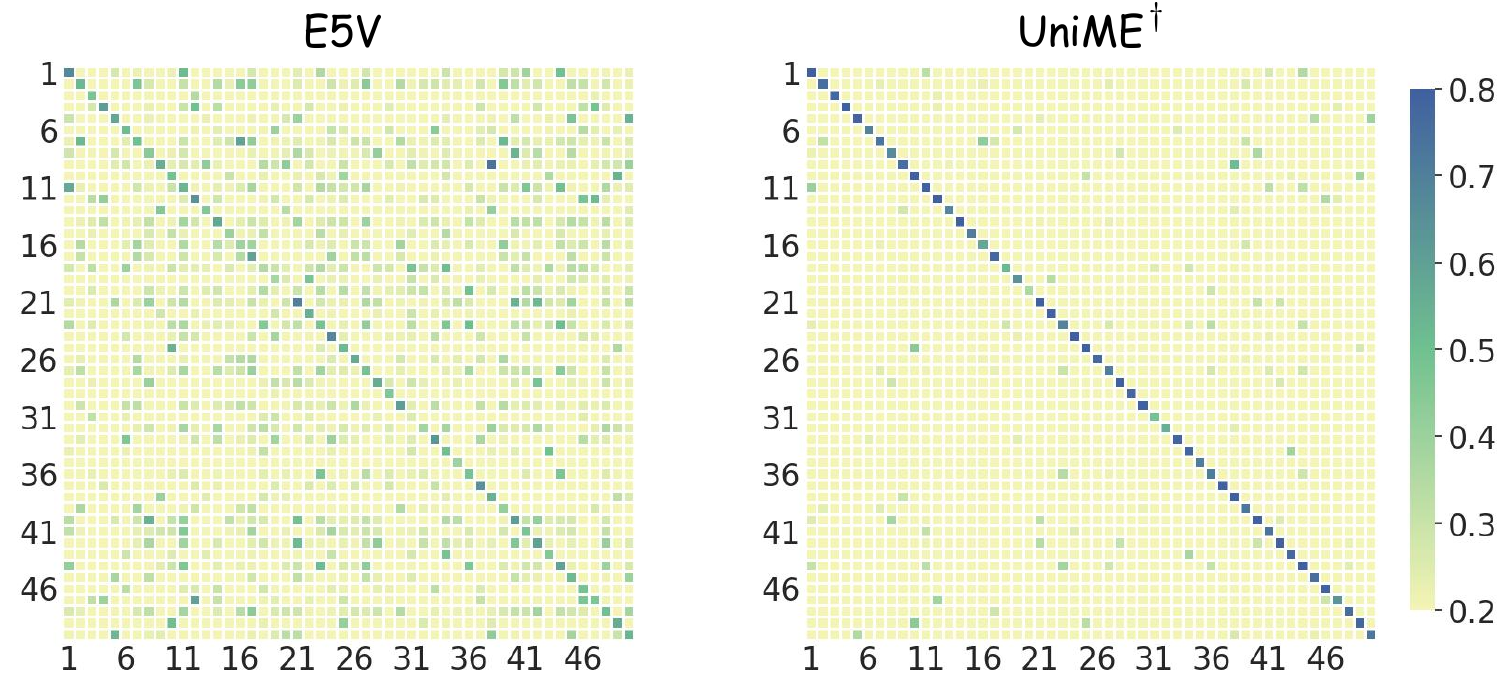}
    \vspace{-5mm}
    \caption{The discrimination comparison between E5-V and UniME$^\dagger$. $^\dagger$ represents the UniME model only training on the first textual discrimination knowledge distillation stage.}
    \Description[<short description>]{<long description>}
    \vspace{-5mm} 
    \label{fig:discrimination}
\end{figure}

In Table~\ref{tab:main_results}, we present the performance of our proposed UniME against existing baseline models. Under identical training data and configuration settings, our proposed UniME consistently achieves significant performance improvements over E5-V across different foundation models. Specifically, UniME exhibits an average performance improvement of 4.2\% over E5-V using the Phi3.5-V model. With LLaVA-1.6 as the base model, UniME further achieves an average enhancement of 4.1\%. These significant performance improvements are primarily due to our proposed textual discrimination knowledge distillation method, which more efficiently enhances the discriminative capability of embeddings from a powerful teacher model. As depicted in Figure~\ref{fig:discrimination}, we randomly select 50 samples from COCO and visualize the cross-modal cosine similarity matrix. The diagonal clarity of the UniME matrix is significantly enhanced compared to that of E5-V, indicating that UniME learns representations with superior distinctiveness. After the hard negative enhanced instruction tuning stage, beneficial from the hard negatives, the discriminative capability of the embeddings from UniME is further improved. Compared with VLM2Vec~\cite{VLM2Vec}, our model achieves 1.3\% and 10.3\% performance improvement when using Phi3.5-V and LLaVA-1.6 as the base model. 

\subsubsection{Short\&Long Caption Cross-Modal Retrieval.} 
We evaluate our approach on zero-shot cross-modal retrieval tasks. Firstly, we conduct experiments on the short-caption datasets Flickr30K and MS-COCO. After the textual discrimination knowledge distillation stage, UniME achieves comparable retrieval performance to E5-V. Subsequent hard negative enhanced instruction tuning further boosts UniME's performance, yielding a significant improvement of 5.2\%–11.3\% over VLM2Vec. For long-caption retrieval tasks on the ShareGPT4V and Urban1K datasets, UniME demonstrates superior performance across all metrics. Specifically, following the textual discriminative knowledge distillation stage, UniME exhibits a performance improvement of 1.3\%-3.8\% based on the Phi3.5-V model. Subsequent enhancement through hard negative enhanced instruction tuning leads to additional gains, with UniME outperforming VLM2Vec by 2.0\%-8.3\%. It is noteworthy that, compared to EVA-CLIP(8B), UniME achieves performance improvements of 14.8\% and 18.1\% in long-caption retrieval on the Urban1K dataset. This significant enhancement primarily stems from the limitation of EVA-CLIP(8B), which is constrained by a 77-token length restriction, thereby inhibiting its ability to fully convey the complete semantic information of long captions.

\subsubsection{Compositional Cross-Modal Retrieval.}
We evaluate the capacity of our UniME model to discriminate hard negative samples using the compositional benchmark SugarCrepe. As shown in Table~\ref{tab:zero-shot-retrieval}, UniME consistently delivers superior performance across all evaluated metrics. Following the textual discriminative knowledge distillation phase, the Phi3.5-V-based UniME outperforms E5-V by 2.0\% in relation replacement, 1.0\% in object swapping, and 15.9\% in attribute addition tasks. After the second hard negative enhanced instruction tuning stage, beneficial from the hard negative, the compositional understanding capabilities of UniME are further enhanced and it achieves 3.9\%, 4.2\%, and 9.1\% performance improvement compared with VLM2Vec. Additionally, UniME exhibits improvements of 4.2\%, 0.6\%, and 6.6\% compared to EVA-CLIP(8B), underscoring its robust capability to discriminate against hard negative samples.

\section{Analysis} 
\label{sec:analysis}

\subsection{Analysis of the Hard Negative}
\begin{figure}[t!]
    \centering
    \includegraphics[width=\linewidth]{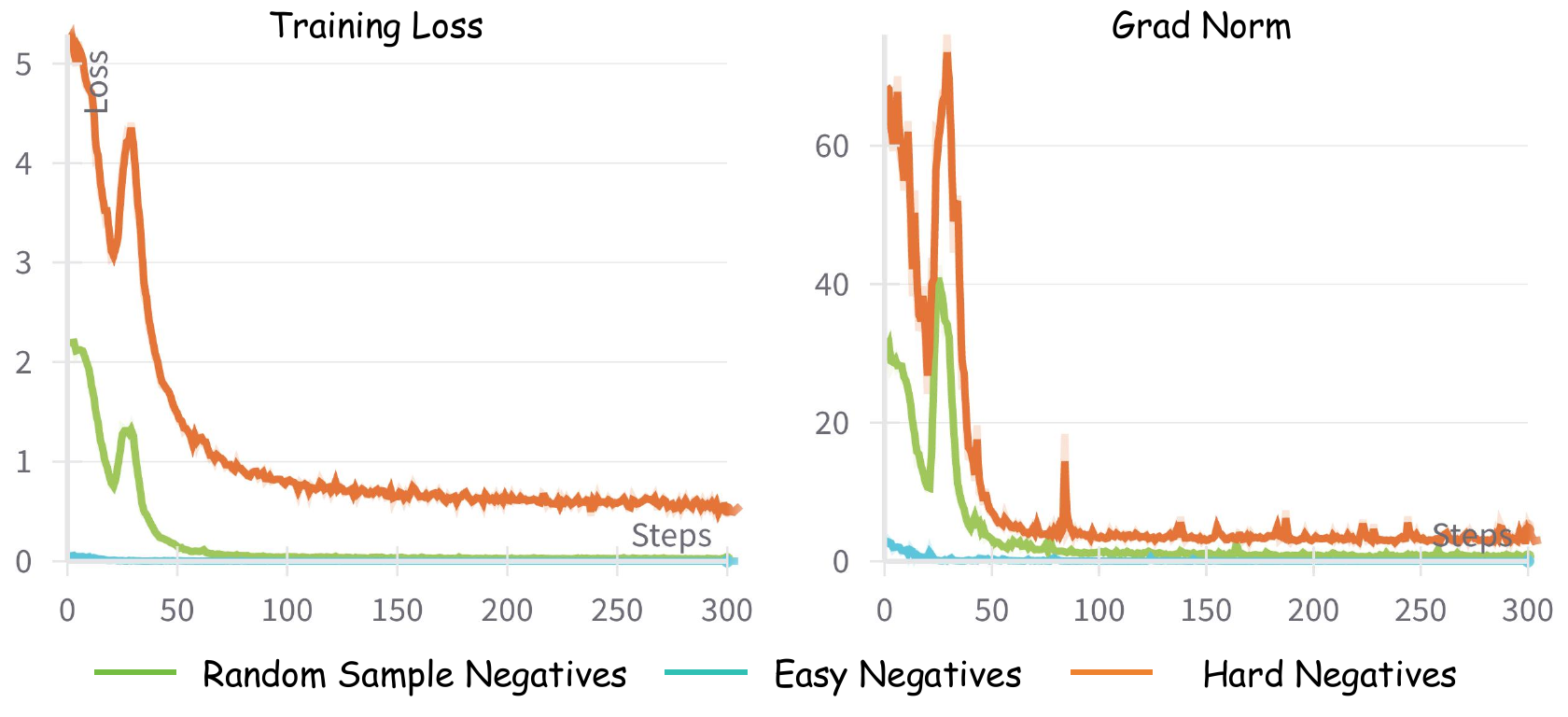}
     \vspace{-5mm}
    \caption{The comparison of training loss and pre-clip gradient norms for hard negatives, easy negatives, and random sample negatives.}
    \Description[<short description>]{<long description>}
    \label{fig:loss_curve_analysis}
     \vspace{-3mm}
\end{figure}

In Figure~\ref{fig:loss_curve_analysis}, we visualize the training loss and pre-clip gradient norms for three negative types: easy negatives (least similar in batch), random negatives (randomly sampled in batch), and hard negatives (most similar in batch after removing positives and false negatives). Since easy negatives are easily distinguishable, the model struggles to enhance its discriminative power through learning from such data, consequently leading to a rapid convergence of the training loss to nearly zero. Using random negatives, the training loss converges more slowly compared to easy negatives, but it eventually nears zero. In contrast, hard negatives pose considerable challenges, sustaining elevated training losses. Correspondingly, gradient norms for easy negatives are minimal, whereas those for hard negatives are substantially higher, differing by orders of magnitude.

\begin{figure*}[t!]
    \centering
    \includegraphics[width=\linewidth]{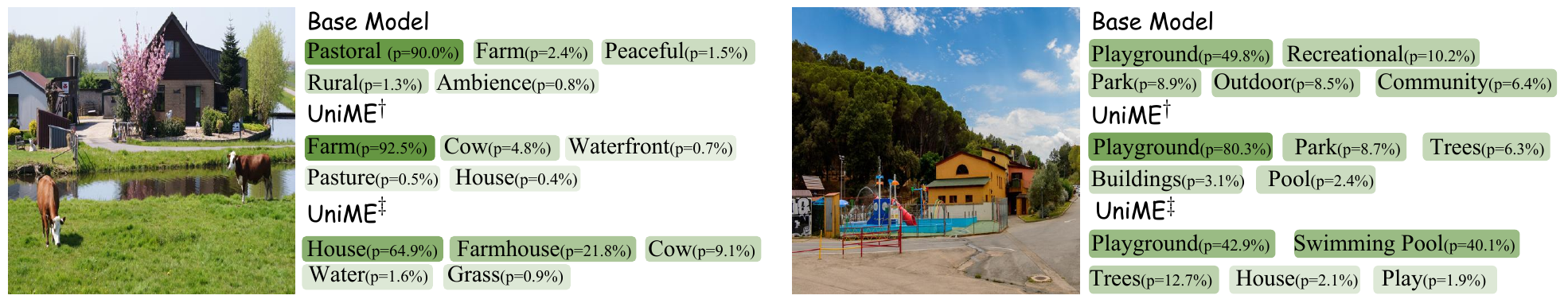}
    \vspace{-7mm}
    \caption{Visualization of the top-k next predicted tokens before and after different training stages based on Phi3.5-V. $^\dagger$: UniME with textual discrimination distillation only. $^\ddagger$: UniME with both textual discrimination distillation and hard negative enhanced instruction tuning.}
    \Description[<short description>]{<long description>}
    \label{fig:visualization}
    \vspace{-2mm}
\end{figure*}

\begin{table}[t]
\centering
\caption{Ablation study of different training stages. We report the mean scores on the MMEB benchmark, short and long cross-modal retrieval, as well as compositional cross-modal retrieval.}
 \vspace{-2mm}
\setlength\tabcolsep{8pt}
\renewcommand\arraystretch{1.2}
\small
\fontsize{8pt}{8pt}\selectfont
\begin{tabular}{ccccccc}
\toprule
 Stage1 & Stage2 & MMEB & $\text{R}_{\text{Short}}$ & $\text{R}_{\text{Long}}$ & $\text{R}_{\text{Compos}}$    \\
 \midrule
 \ding{56} & \ding{56}  & 25.3  & 44.2 & 62.9 & 63.1 \\
 \ding{52} & \ding{56} &  40.3 & 63.7 & 87.8 & 83.0  \\
 \ding{56} & \ding{52} & 63.8 & 61.5 & 84.2 & 77.1  \\
 \ding{52} & \ding{52} & \textbf{64.2} & \textbf{70.4} & \textbf{94.1} & \textbf{84.8}  \\

\bottomrule
\end{tabular}
\label{tab:ablation_stage}
\vspace{-3mm}
\end{table}

\subsection{Ablation on Training Stages}
We conduct an ablation study based on Phi3.5-V to evaluate different training stages. As shown in Table~\ref{tab:ablation_stage}, the initial embeddings from Phi3.5-V exhibit weak discriminative properties, leading to suboptimal task performance. After the initial stage of textual discriminative knowledge distillation, the model registers performance improvements of 15\%, 19.5\%, 24.9\%, and 19.9\% on the MMEB benchmark, short and long caption cross-modal retrieval, and compositional retrieval tasks, respectively. Focusing solely on the second stage, which involves hard negative enhanced instruction tuning, results in performance gains of 38.5\%, 17.3\%, 21.3\%, and 14.0\% in the same tasks. Notably, the enhancement in MMEB benchmark performance after the second stage markedly exceeds that of the first, primarily due to improved model capabilities in following complex instructions. By integrating both training stages, our UniME model achieves optimal performance across all evaluated downstream tasks.

\subsection{Ablation on the Threshold \texorpdfstring{$\beta$}{Lg}}
The threshold $\beta$ directly influences the percentage of the filtered false negatives. We conduct an ablation study on 20\% of the entire dataset to identify the optimal value of $\beta$. As illustrated in Table~\ref{tab:alpha}, setting $\beta$ to -0.1 results in filtering false negatives for 81.7\% of samples. However, the inclusion of some hard negative samples in the filtered set results in poor model performance. Increasing $\beta$ from -0.1 to 0.1 reduces the proportion of samples with filtered false negatives from 81.7\% to 22.9\%, thereby significantly improving performance. Further increasing $\beta$ to 0.3 filters false negatives in only 13.1\% of samples, resulting in a slight performance decline due to persistent false negatives.

\begin{table}[t]
\centering
\caption{Ablation study of the false negative filtering threshold $\beta$. FalseNeg(\%): proportion of samples which filtered false negatives.}
\setlength\tabcolsep{7pt}
\renewcommand\arraystretch{1.2}
\small
\fontsize{8pt}{8pt}\selectfont
\begin{tabular}{lccccc}
\toprule
\multicolumn{1}{c}{\multirow{2}{*}{\textbf{Model}}} & \multicolumn{1}{c}{\multirow{2}{*}{\textbf{$\beta$}}} & \multicolumn{1}{c}{\multirow{2}{*}{\textbf{FalseNeg(\%)}}}  & \multicolumn{3}{c}{\textbf{Average Score}}    \\ \cmidrule(lr){4-6} 
\multicolumn{1}{c}{} & \multicolumn{1}{c}{} &  \multicolumn{1}{c}{} & IND & OOD & Overall \\ \midrule
\multirow{5}{*}{Phi3.5-V} & -0.1 & 81.7\% & 61.0 & 43.4 & 55.3 \\
 & 0.0 & 53.2\%  & 66.1 & 49.0 & 61.1 \\
 & 0.1 & 22.9\%  & 68.2 & \textbf{52.7} & \textbf{64.2} \\
 & 0.2 & 18.2\%  & 68.2 & 51.9 & 63.7 \\
 & 0.3 & 13.1\%  & \textbf{68.3} & 52.1 & 63.9 \\

\bottomrule
\end{tabular}
\vspace{-3mm} 
\label{tab:alpha}
\end{table}

\subsection{Ablation on the Number of Hard Negatives}
The value of $k$ directly determines the number of hard samples sampled for each instance in a batch. In Table~\ref{tab:ablation_k}, we illustrates the impact of ($k$) on the performance of UniME based on Phi-3.5V. When we set $k$=8, compared to $k$=4, the higher diversity of hard negative samples leads to superior performance. Further increasing the value of $k$ introduces easy negatives, causing the model to lose focus on learning hard negatives, gradually reducing performance.

\subsection{Visualization of the Output Distribution}
To further explore the semantic representations captured by UniME embeddings, we utilize the prompt \textit{"<Image> Summary above image in one word: $\backslash$n"} and visualize the prediction probability of the top-k next predicted tokens before and after our proposed different training stages in Figure~\ref{fig:visualization}. We observe that before training, the predicted tokens are more abstract, such as ``Pastoral'' and ``Peaceful''. After the textual discriminative knowledge distillation, the tokens shift towards more concrete semantics, including ``cow'', ``waterfront'', and ``house'', though the probability distribution remains largely concentrated on "Farm". After the second stage of hard negative enhanced instruction tuning, the probability distribution becomes more evenly spread across multiple tokens that align with the image's semantics, thereby allowing the embeddings to more accurately express the semantic content of the image with enhanced discriminative capability.

\begin{table}[t!]
\centering
\caption{Ablation study of the number $k$ of the hard negatives. HardNeg(\%): proportion of hard negative samples within a batch.}
\vspace{-1mm}
\setlength\tabcolsep{6pt}
\renewcommand\arraystretch{1.2}
\small
\fontsize{8pt}{8pt}\selectfont
\begin{tabular}{lccccc}
\toprule
\multicolumn{1}{c}{\multirow{2}{*}{\textbf{Model}}} & \multicolumn{1}{c}{\multirow{2}{*} {\textbf{Top-$k$}}} &  
\multicolumn{1}{c}{\multirow{2}{*}{\textbf{HardNeg(\%)}}}& \multicolumn{3}{c}{\textbf{Average Score}}    \\ 
\cmidrule(lr){4-6} 
\multicolumn{1}{c}{} & \multicolumn{1}{c}{} & \multicolumn{1}{c}{} & IND & OOD & Overall \\ \midrule
\multirow{5}{*}{Phi3.5-V}  & 4  & 0.4\% & 67.8 & 52.4 & 63.8 \\
& 8  & 0.8\% & 68.2 & \textbf{52.7} & \textbf{64.2} \\
& 16 & 1.6\% & 68.1 & 51.6 & 63.5 \\
& 32 & 3.2\% & \textbf{68.4} & 51.8 & 63.6 \\
& 64 & 6.4\% & 68.1 & 51.1 & 63.2 \\
\bottomrule
\end{tabular}
\vspace{-5mm}
\label{tab:ablation_k}
\end{table}
\section{Conclusion}
In this paper, we introduce \textbf{UniME}~(\textbf{Uni}versal \textbf{M}ultimodal \textbf{E}mbedding), a novel two-stage framework that enables large multimodal language models with the capacity to learn discriminative representations applicable to a variety of downstream tasks. In the first textual discriminative knowledge distillation stage, we leverage a powerful LLM-based teacher model to enhance the embedding capability of the MLLM’s language component. In the second hard negative enhanced instruction tuning stage, we initially mitigate false negative contamination and then sample multiple hard negatives per instance within each batch, forcing the model to focus on challenging samples. This approach not only improves discriminative power but also enhances instruction-following ability in downstream tasks. We conduct extensive experiments on the MMEB benchmark and multiple retrieval tasks, including short\&long caption retrieval and compositional retrieval. Results demonstrate that UniME achieves consistent performance improvement across all tasks, exhibiting superior discriminative and compositional capabilities. We hope that our work provides insights into multimodal representation learning.

\bibliographystyle{ACM-Reference-Format}
\balance
\bibliography{sample-base}

\clearpage
\appendix

This supplementary material provides additional details on our experimental settings, including training configurations and evaluation benchmarks as described in Sec.\ref{sec:deatiled experiemnt settings}. Additionally, it presents expanded results in Sec.~\ref{sec:external_results}, including an ablation study on the LoRA rank and detailed findings on the MMEB benchmark. Finally, Sec.~\ref{sec:further_analysis} provides supplementary visualizations that depict the output distribution and examples of negative data.

\begin{table}[h!]
\centering
\caption{Training hyperparameters and computational requirements for \mdname(Phi3.5-V) and \mdname(LLaVA-1.6).}
\label{tab:training_details}
\begin{tabular}{@{}lcc@{}}
\toprule
\multicolumn{1}{c}{\textbf{Hyperparameter}} & \textbf{\mdname(Phi3.5-V)} & \textbf{\mdname(LLaVA-1.6)} \\
\midrule
\multicolumn{3}{c}{\textbf{\textit{Stage 1:Textual Discriminative Knowledge Distillation}}} \\
\midrule
Training samples & \multicolumn{2}{c}{273K} \\
Batch size & \multicolumn{2}{c}{768} \\
Learning rate & \multicolumn{2}{c}{5$\times$10$^{-4}$} \\
LoRA rank & \multicolumn{2}{c}{32} \\
Epochs & \multicolumn{2}{c}{2} \\
DeepSpeed stage & \multicolumn{2}{c}{2} \\
GPU configuration & \multicolumn{2}{c}{8$\times$A100} \\
Precision & \multicolumn{2}{c}{FP16} \\
Training time & 1 hour & 2 hours \\
\midrule
\multicolumn{3}{c}{\textbf{\textit{Stage 2: Hard Negative Enhanced Instruction Tuning}}} \\
\midrule
Training samples & \multicolumn{2}{c}{662K} \\
Batch size & \multicolumn{2}{c}{1024} \\
Learning rate & 1$\times$10$^{-4}$ & 2$\times$10$^{-5}$ \\
LoRA rank & \multicolumn{2}{c}{16} \\
Training steps & \multicolumn{2}{c}{1000} \\
Optimizer & \multicolumn{2}{c}{AdamW} \\
GPU configuration & \multicolumn{2}{c}{8$\times$A100} \\
Precision & \multicolumn{2}{c}{BF16} \\
Training time & 26 hours & 37 hours \\
\bottomrule
\end{tabular}
\label{tab:supp_training_parameters}
\end{table}

\section{Detailed Experiment Settings}
\label{sec:deatiled experiemnt settings}

\subsection{Training Details}

We provide the training configurations of \mdname\ in Table~\ref{tab:supp_training_parameters}.

\noindent{\bf Stage1: Textual Discriminative Knowledge Distillation.}
We use 8$\times$A100 GPUs~(80GB each) to train \mdname\ with text-only data. The text embeddings of the teacher model are extracted offline, and the model is trained by utilizing QLoRA~\cite{QLora}, training durations are significantly reduced. Specifically, training completes in one hour for the Phi3.5-V~\cite{phi3} backbone and two hours for the LLaVA-1.6~\cite{llavanext} backbone.

\noindent{\bf Stage2: Hard Negative Enhanced Instruction Tuning.} 
Upon integrating the backbone with stage 1 LoRA outputs, we further advance \mdname\ on image-text pairs derived from the MMEB training set. Under identical hardware conditions, this phase requires 26 hours for Phi-3.5V and 37 hours for LLaVA-1.6.

\subsection{Retrieval Task Evaluation Benchmarks}
We evaluate \mdname\ on diverse retrieval benchmarks, including short-caption, long-caption, and compositional image-text tasks~(as shown in Table~\ref{tab:supp_evaluation_benchmark}). For each benchmark, we adopt its standard evaluation protocol. In retrieval tasks, we primarily report Recall@1 as the evaluation metric.
\begin{table}[h!]
\centering
\tiny 
\caption{\textbf{Summary of the evaluation benchmarks.} \# Queries represents the number of test queries, and \# Candidates denotes the number of test candidates per query.
}
\resizebox{.47\textwidth}{!}{ 
\setlength{\tabcolsep}{1mm}{
  \begin{tabular}{lccc}
    \toprule
    \textbf{Benchmark} & \textbf{Zero-shot} & \textbf{\#Queries} & \textbf{\#Candidates} \\
    \midrule
    Flickr30K~\cite{flickr30k} & \ding{52} & 1K & 5K \\
    COCO~\cite{MSCOCO2014} & \ding{52} & 5K & 25K \\
    ShareGPT4V~\cite{sharegpt4v} & \ding{52} & 1K & 1K\\
    Urban1K~\cite{longclip} & \ding{52} & 1K & 1K\\
    SugarCrepe~\cite{sugarcrepe} & \ding{52} & 7.5K & 2 \\
    \bottomrule
  \end{tabular}
}
}
\vspace{-2mm}
\label{tab:supp_evaluation_benchmark}
\end{table}

\section{External Results}
\label{sec:external_results}

\begin{table}[ht]
\centering
\caption{Performance comparison of different LoRA ranks under two training stages (IND: In-Domain, OOD: Out-Of-Domain)}
\label{tab:alpha}
\renewcommand{\arraystretch}{1.25}
\setlength{\tabcolsep}{6pt}
\begin{tabular}{lcccc}
\toprule
\multicolumn{1}{c}{\multirow{2}{*}{\textbf{Stage}}} & \multirow{2}{*}{\textbf{LoRA}} & \multicolumn{3}{c}{\textbf{Average Score}} \\
\cmidrule(lr){3-5}
 & & \multicolumn{1}{c}{\textbf{IND}} & \multicolumn{1}{c}{\textbf{OOD}} & \multicolumn{1}{c}{\textbf{Overall}} \\
\midrule
\multirow{5}{*}{Stage1} & 4 & 36.0 & 37.9 & 40.0 \\
 & 8 & 35.8 & 37.8 & 39.9 \\
 & 16 & 35.9 & 38.0 & 40.0 \\
 & 32 & \textbf{36.0} & \textbf{38.3} & \textbf{40.3} \\
 & 64 & 35.6 & 37.8 & 39.9 \\
 
\midrule
\multirow{5}{*}{Stage2} & 4 & 67.4 & 51.0 & 62.6 \\
 & 8 & 68.2 & 51.0 & 63.8 \\
 & 16 & \textbf{68.2} & \textbf{52.7} & \textbf{64.2} \\
 & 32 & 67.6 & 51.8 & 63.2 \\
 & 64 & 67.0 & 51.5 & 62.7 \\

\bottomrule
\end{tabular}
\label{tab:supp_ablation_of_lora}
\vspace{-3mm}
\end{table}

\begin{figure*}[t!]
    \centering
    \includegraphics[width=\linewidth]{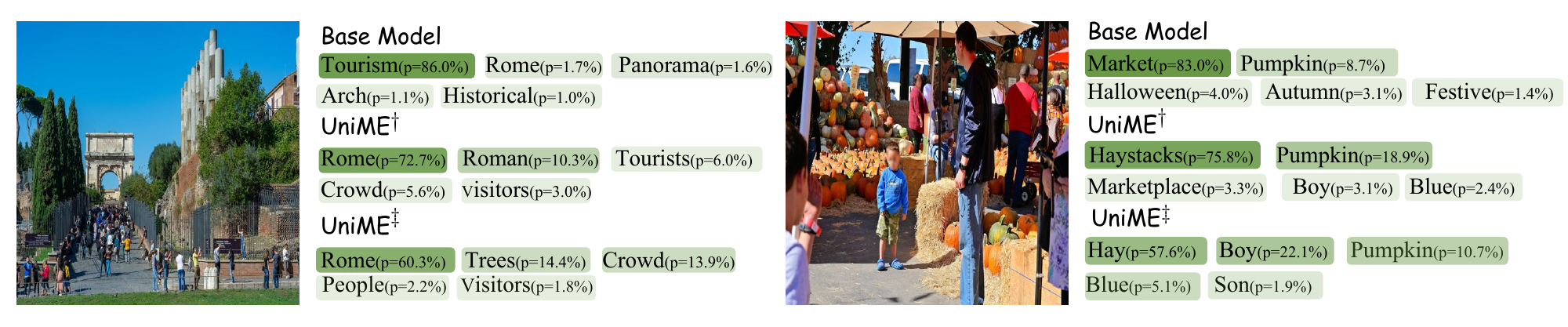}
    \caption{Visualization of the top-k next predicted tokens before and after different training stages based on Phi3.5-V. $^\dagger$: UniME with textual discrimination distillation only. $^\ddagger$: UniME with both textual discrimination distillation and hard negative enhanced instruction tuning.}
    \Description[<short description>]{<long description>}
    \label{fig:supp_visualization}
\end{figure*}

\subsection{Ablation on the LoRA Rank}
\sloppy
Due to computational limitations and the proven efficacy of LoRA~\cite{QLora}, we utilize QLoRA to fine-tune the backbone of \mdname. As detailed in Table~\ref{tab:supp_ablation_of_lora}, we assess the impact of various LoRA ranks on the MMEB benchmarks~\cite{VLM2Vec} utilizing the Phi-3.5V backbone. Our findings indicate that a LoRA rank of 32 delivers optimal performance in Stage 1, whereas a rank of 16 is most effective in Stage 2.

\subsection{Specific results on the MMEB}
We present comparative results of eight models on the MMEB benchmark in Table~\ref{tab:supp_main_MMEB_exp_per_task}. The performance metrics for CLIP, SigLIP, BLIP-2, and MagicLens are sourced directly from VLM2Vec~\cite{VLM2Vec}. Conversely, results for EVA-CLIP(8B), E5-V, VLM2Vec, and UniME are obtained through reproduction in our experiments. For E5-V, VLM2Vec, and UniME, we only report the metrics for their best-performing variants, all of which employ the LLaVA-1.6~\cite{llavanext} backbone.

\begin{figure}[t!]
    \centering
    \includegraphics[width=\linewidth]{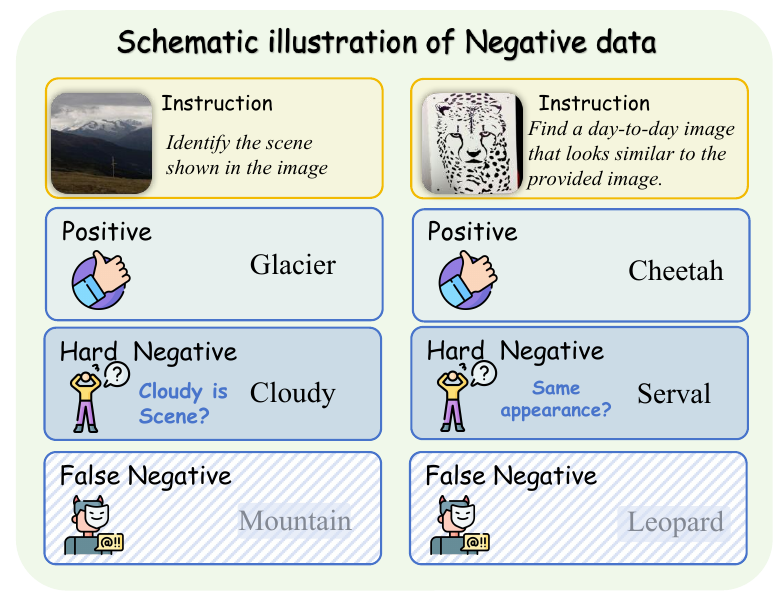}
    \caption{Schematic illustration of Negative data.}
    \Description[<short description>]{<long description>}
    \label{fig:supp_hard_negative}
\end{figure}

\section{Further Analysis}
\label{sec:further_analysis}

\begin{table*}[ht]
\centering
\caption{The detailed results of the baselines and our \mdname on MMEB, which includes 20 in-distribution datasets and 16 out-of-distribution datasets. The out-of-distribution datasets are highlighted with a yellow background in the table. We only introduce the best version of \mdname, E5-V and VLM2Vec in the table, which uses LLaVA-1.6 as the backbone.}
\resizebox{0.9\textwidth}{!}{
\begin{tabular}{lcccccccc}
\toprule
\rowcolor{gray!30}  & \textbf{CLIP} & \textbf{SigLIP} & \textbf{BLIP2} & \textbf{MagicLens} & \textbf{EVA-CLIP} & \textbf{E5-V} & \textbf{VLM2Vec} & \textbf{\mdname} \\
\midrule
\rowcolor{orange!30} \textbf{Classification (10 tasks)} & & & & & & & & \\
ImageNet-1K          & 55.8 & 45.4 & 10.3 & 48.0 & 75.0 & 40.5 & 66.5 & 71.3 \\
N24News              & 34.7 & 13.9 & 36.0 & 33.7 & 33.8 & 31.5 & 76.4 & 79.5 \\
HatefulMemes         & 51.1 & 47.2 & 49.6 & 49.0 & 49.3 & 49.3 & 60.9 & 64.6 \\
VOC2007              & 50.7 & 64.3 & 52.1 & 51.6 & 44.3 & 76.7 & 84.0 & 90.4 \\
SUN397               & 43.4 & 39.6 & 34.5 & 57.0 & 62.7 & 52.3 & 73.2 & 75.9 \\
\rowcolor{yellow!15} Place365  & 28.5 & 20.0 & 21.5 & 31.5 & 38.7 & 32.0 & 42.1 & 45.6 \\
\rowcolor{yellow!15} ImageNet-A & 25.5 & 42.6 & 3.2  & 8.0  & 54.8 & 18.2 & 39.9 & 45.5 \\
\rowcolor{yellow!15} ImageNet-R & 75.6 & 75.0 & 39.7 & 70.9 & 95.4 & 56.7 & 74.6 & 78.4 \\
\rowcolor{yellow!15} ObjectNet  & 43.4 & 40.3 & 20.6 & 31.6 & 67.8 & 34.2 & 34.3 & 36.4 \\
\rowcolor{yellow!15} Country-211 & 19.2 & 14.2 & 2.5  & 6.2  & 38.7 & 5.9 & 16.1 & 18.7 \\
\textit{All Classification} & 42.8 & 40.3 & 27.0 & 38.8 & 56.0 & 39.7 & 56.8 & 60.6 \\
\midrule

\rowcolor{blue!30} \textbf{VQA (10 tasks)} & & & & & & & & \\
OK-VQA               & 7.5  & 2.4  & 8.7  & 12.7 & 9.9 & 15.1 & 66.5 & 68.3 \\
A-OKVQA              & 3.8  & 1.5  & 3.2  & 2.9  & 2.8 & 4.7 & 54.9 & 58.7 \\
DocVQA               & 4.0  & 4.2  & 2.6  & 3.0  & 7.4 & 9.1 & 64.4 & 67.6 \\
InfographicsVQA      & 4.6  & 2.7  & 2.0  & 5.9  & 6.0 & 8.7 & 34.8 & 37.0 \\
ChartQA              & 1.4  & 3.0  & 0.5  & 0.9  & 1.5 & 4.2 & 33.1 & 33.4 \\
Visual7W             & 4.0  & 1.2  & 1.3  & 2.5  & 2.2 & 4.5 & 49.8 & 51.7 \\
\rowcolor{yellow!15} ScienceQA  & 9.4  & 7.9  & 6.8  & 5.2  & 14.1 & 9.6 & 37.3 & 40.5 \\
\rowcolor{yellow!15} VizWiz    & 8.2  & 2.3  & 4.0  & 1.7  & 4.3 & 8.6 & 39.9 & 42.7 \\
\rowcolor{yellow!15} GQA        & 41.3 & 57.5 & 9.7  & 43.5 & 44.7 & 34.1 & 57.3 & 63.6 \\
\rowcolor{yellow!15} TextVQA    & 7.0  & 1.0  & 3.3  & 4.6  & 10.8 & 9.5 & 65.7 & 65.2 \\
\textit{All VQA}      & 9.1  & 8.4  & 4.2  & 8.3  & 10.4 & 10.8 & 50.4 & 52.9 \\
\midrule

\rowcolor{green!30} \textbf{Retrieval (12 tasks)} & & & & & & & & \\
VisDial              & 30.7 & 21.5 & 18.0 & 24.8 & 20.4 & 57.6 & 75.3 & 79.7 \\
CIRR                 & 12.6 & 15.1 & 9.8  & 39.1 & 36.0 & 41.0 & 51.3 & 52.2 \\
VisualNews\_t2i      & 78.9 & 51.0 & 48.1 & 50.7 & 82.4 & 43.9 & 70.7 & 74.8 \\
VisualNews\_i2t      & 79.6 & 52.4 & 13.5 & 21.1 & 88.2 & 46.8 & 75.2 & 78.8 \\
MSCOCO\_t2i          & 59.5 & 58.3 & 53.7 & 54.1 & 65.3 & 68.6 & 69.9 & 74.9 \\
MSCOCO\_i2t          & 57.7 & 55.0 & 20.3 & 40.0 & 67.2 & 54.8 & 67.7 & 73.8 \\
NIGHTS               & 60.4 & 62.9 & 56.5 & 58.1 & 0.2 & 0.1 & 63.3 & 66.2 \\
WebQA                & 67.5 & 58.1 & 55.4 & 43.0 & 70.9 & 33.7 & 83.6 & 89.8 \\
\rowcolor{yellow!15} FashionIQ  & 11.4 & 20.1 & 9.3  & 11.2 & 16.1 & 11.2 & 15.2 & 16.5 \\
\rowcolor{yellow!15} Wiki-SS-NQ & 55.0 & 55.1 & 28.7 & 18.7 & 46.7 & 61.0 & 63.4 & 66.6 \\
\rowcolor{yellow!15} OVEN       & 41.1 & 56.0 & 39.5 & 1.6  & 1.8 & 0.5 & 49.6 & 55.7 \\
\rowcolor{yellow!15} EDIS       & 81.0 & 23.6 & 54.4 & 62.6 & 95.6 & 53.8 & 73.7 & 86.2 \\
\textit{All Retrieval} & 53.0 & 31.6 & 33.9 & 35.4 & 49.2 & 39.4 & 63.3 & 67.9 \\
\midrule

\rowcolor{purple!30} \textbf{Visual Grounding (4 tasks)} & & & & & & & & \\
MSCOCO         & 33.8 & 46.4 & 28.9 & 22.1 & 35.8 & 41.7 & 77.0 & 76.5 \\
\rowcolor{yellow!15} RefCOCO  & 56.9 & 70.8 & 47.4 & 22.8 & 59.9 & 62.2 & 85.9 & 89.3 \\
\rowcolor{yellow!15} RefCOCO-matching  & 61.3 & 50.8 & 59.5 & 35.6 & 70.0 & 74.9 & 83.8 & 90.6 \\
\rowcolor{yellow!15} Visual7W-pointing & 55.1 & 70.1 & 52.0 & 23.4 & 70.2 & 61.8 & 83.6 & 84.1 \\
\textit{All Visual Grounding} & 51.8 & 59.5 & 47.0 & 26.0 & 58.9 & 60.2 & 82.6 & 85.1 \\
\midrule

\rowcolor{cyan!15} \textbf{Final Score (36 tasks)} & & & & & & & & \\
All        & 39.2 & 35.0 & 28.0 & 27.1 & 43.7 & 37.5 & 63.3 & 66.6 \\
All IND    & 37.1 & 32.3 & 25.3 & 31.0 & 38.1 & 34.2 & 64.9 & 68.4 \\
All OOD    & 38.7 & 38.0 & 25.1 & 23.7 & 45.6 & 33.4 & 53.9 & 57.9 \\

\bottomrule
\end{tabular}
}
\label{tab:supp_main_MMEB_exp_per_task}
\end{table*}

\subsection{Visualization of the Output Distribution}
In Figure~\ref{fig:supp_visualization}, we further present additional examples that compare the top-k prediction probabilities of subsequent tokens, illustrating the evolution of model behavior across various training stages of our UniME.

\subsection{Schematic illustration of Negative Data}
Figure~\ref{fig:supp_hard_negative} provides additional examples of negative samples encountered during training. Our observations indicate that false negatives are more prevalent in textual data than in visual data, attributable primarily to the presence of synonyms.

\end{document}